\documentclass[preprint,journal]{vgtc}       

\ifpdf
  \pdfoutput=1\relax                   
  \usepackage{graphicx}                
  \DeclareGraphicsExtensions{.pdf,.png,.jpg,.jpeg} 
\else
  \usepackage{graphicx}                
  \DeclareGraphicsExtensions{.eps}     
\fi%

\graphicspath{{figures/}{pictures/}{images/}{./}} 

\usepackage{microtype}                 
\PassOptionsToPackage{warn}{textcomp}  
\usepackage{textcomp}                  
\usepackage{mathptmx}                  
\usepackage{times}                     
\usepackage{cite}                      
\usepackage{tabu}                      
\usepackage{booktabs}                  
\usepackage{wrapfig}
\usepackage{xcolor}

\definecolor{c1}{rgb}{0,0.3,1}
\definecolor{c2}{rgb}{0.4,0,0.7}

\newcommand{\revision}[1]{\textcolor{black}{#1}}

\definecolor{pink}{RGB}{233,119,130}
\definecolor{yellow}{RGB}{235,199,68}
\definecolor{green}{RGB}{128,150,72}
\definecolor{blue}{RGB}{29,75,105}

\definecolor{bluePixel}{RGB}{82,182,189}
\definecolor{orangePixel}{RGB}{249,111,10}

\usepackage{microtype}     

\usepackage[export]{adjustbox}
\definecolor{light-gray}{gray}{0.95}

\onlineid{0}

\vgtccategory{Research}
\vgtcpapertype{please specify}

\title{Visual Comparison of Language Model Adaptation}

\author{Rita Sevastjanova, Eren Cakmak, Shauli Ravfogel, Ryan Cotterell, and Mennatallah El-Assady}
\authorfooter{
\item
 Rita Sevastjanova and Eren Cakmak are with  University of Konstanz. E-mail: firstname.lastname@uni-konstanz.de.
 \item
Shauli Ravfogel is with Bar-Ilan University. E-mail: shauli.ravfogel@gmail.com.
 \item
Ryan Cotterell is with ETH. E-mail: ryan.cotterell@inf.ethz.ch.
\item
 Mennatallah El-Assady is with ETH, AI Center. E-mail: melassady@ethz.ch.
}

\shortauthortitle{Biv \MakeLowercase{\textit{et al.}}: Global Illumination for Fun and Profit}

\abstract{Neural language models are widely used; 
however, their model parameters often need to be adapted to the specific domains and tasks of an application, which is time- and resource-consuming.
Thus, adapters have recently been introduced as a lightweight alternative for model adaptation. They consist of a small set of task-specific parameters with a reduced training time and simple parameter composition. 
The simplicity of adapter training and composition comes along with new challenges, such as maintaining an overview of adapter properties and effectively comparing their produced embedding spaces.
To help developers overcome these challenges, we provide a twofold contribution.
First, in close collaboration with NLP researchers, we conducted a requirement analysis for an approach supporting adapter evaluation and detected, among others, the need for both intrinsic (i.e., embedding similarity-based) and extrinsic (i.e., prediction-based) explanation methods.
Second, motivated by the gathered requirements, we designed a flexible visual analytics workspace that enables the comparison of adapter properties. 
In this paper, we discuss several design iterations and alternatives for interactive, comparative visual explanation methods.
Our comparative visualizations show the differences in the adapted embedding vectors and prediction outcomes for diverse human-interpretable concepts (e.g., \textit{person names}, \textit{human \revision{qualities}}). 
\revision{We evaluate our workspace through case studies and show that, for instance, an adapter trained on the language debiasing task according to context-0 (\textit{decontextualized}) embeddings introduces a new type of bias where words (even gender-independent words such as countries) become more similar to female- than male pronouns. We demonstrate that these are artifacts of context-0 embeddings, and the adapter effectively eliminates the gender information from the contextualized word representations. 
}
} 


\CCScatlist{ 
 \CCScat{K.6.1}{Management of Computing and Information Systems}%
{Project and People Management}{Life Cycle};
 \CCScat{K.7.m}{The Computing Profession}{Miscellaneous}{Ethics}
}

\teaser{
  \centering
   \includegraphics[width=\textwidth]{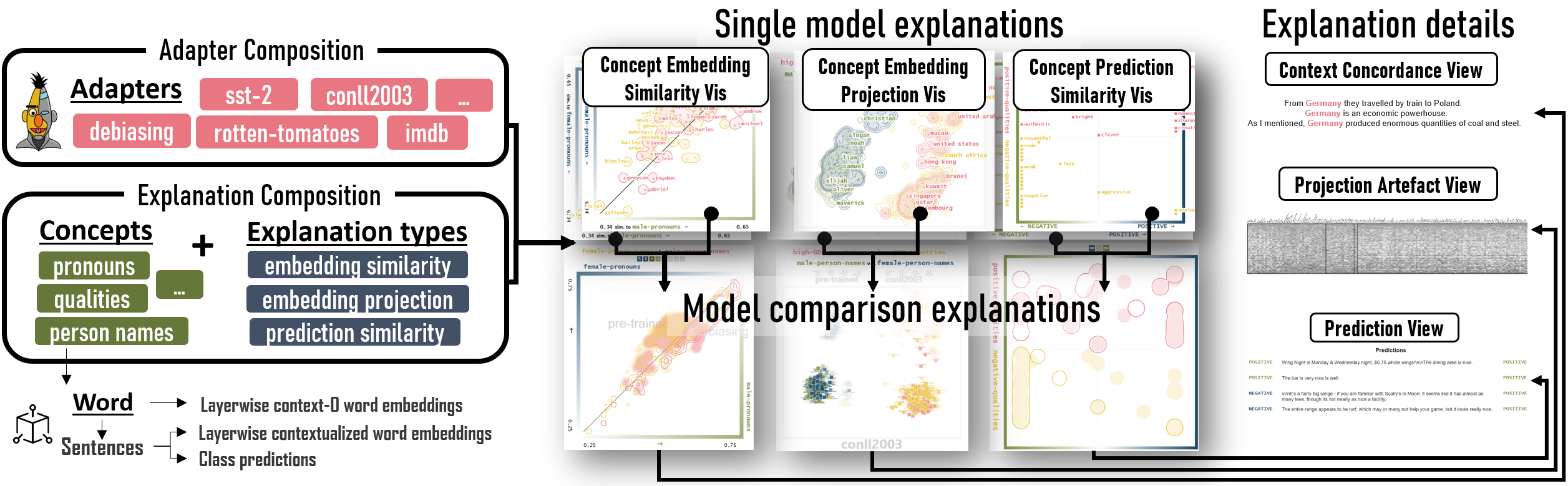}
    \caption{We present a workspace that enables the evaluation and comparison of adapters -- lightweight alternatives for language model fine-tuning. After data pre-processing (e.g., embedding extraction), users can select pre-trained adapters, create explanations, and explore model differences through three types of visualizations: \textit{Concept Embedding Similarity}, \textit{Concept Embedding Projection},
and \textit{Concept Prediction Similarity}. The explanations are provided for single models as well as model comparisons. For each explanation, we provide further explanation details, such as the word contexts as well as embedding vectors themselves. }
  \label{fig:teaser}
}

\vgtcinsertpkg
\usepackage{subcaption}

\begin{document}


\firstsection{Introduction}

\maketitle

Language models (LMs) such as the masked language model BERT~\cite{devlin2018bert} are widely used for diverse natural language processing (NLP) and understanding tasks. 
Such models are capable of learning manifold language properties in an unsupervised manner~\cite{rogers2020primer}. 
However, the model parameters typically need to be updated before using them on downstream tasks, such as sentiment classification.
Task specific fine-tuning~\cite{howard-ruder-2018-universal, qiu2020pre} along with domain specific fine-tuning~\cite{Han2019UnsupervisedDA, gururangan-etal-2020-dont} are the most common methods for parameter adaptation.
Although fine-tuning methods commonly achieve state-of-the-art results on many NLP tasks~\cite{qiu2020pre}, they come along with limitations such as a high training time and storage~\cite{doi:10.1073/pnas.1611835114}. 
To overcome the shortcomings of the model fine-tuning, Houlsby et al.~\cite{houlsby2019parameter} have recently introduced adapter modules -- a lightweight alternative for LM fine-tuning. 
Instead of adapting the complete model, adapters learn a small set of task-specific parameters, requiring less training time and storage space.
For a more efficient adapter training and composition, Pfeiffer et al.~\cite{pfeiffer2020AdapterHub} have proposed a modular adapter framework called AdapterHub. 
It comes along with adapter-transformers -- an extension of HuggingFace's transformers library\footnote{\url{https://github.com/Adapter-Hub/adapter-transformers}}, integrating adapters into state-of-the-art LMs. 
In addition to the simple parameter adaptation, the AdapterHub framework allows sharing adapters with the community, supporting open science practices. 

\revision{The AdapterHub repository currently contains almost 400 adapters for 72 text analysis tasks and 50 languages.
To select the best adapter for a given analysis task, one needs to be able to compare the adapters and their learned language properties. 
The related work has shown that such model comparison tasks are the focus of both model- and data-driven users working with LMs~\cite{boggust2022embedding}.} 
To understand more about the typical analysis setting, data, and performed tasks when evaluating fine-tuned model properties, we \revision{conducted literature review and} semi-structured interviews with two NLP researchers.
The requirement analysis revealed that researchers are interested in analyzing models with respect to different human-interpretable concepts.
In particular, they investigate how specific concept representations change during fine-tuning.
The analysis is typically performed on two types of data: (1) word embedding representations and (2) classifier prediction outcomes. 
Using word embeddings, they analyze evolving concept intersections as well as newly produced artifacts like strange word associations (e.g., biases). 
Prediction outcomes are used to analyze task-adapted model behavior changes, e.g., whether specific word associations lead to unexpected prediction outcomes. 

The adapters trained on one particular task typically have different architectures~\cite{houlsby2019parameter, pfeiffer-etal-2020-mad} and training corpora.
These different learning settings usually lead to different model performances; it is difficult, though, to keep track of such performance variations. 
The continuous development of new adapters thus dictates the need for a solution that assists the analysis and comparison of adapter properties.

To support the NLP community in an effective adapter evaluation and comparison, we contribute a novel visual analytics workspace.
The workspace integrates adapters from the AdapterHub repository and enables their analysis through three types of visual explanation methods: \textit{Concept Embedding Similarity}, \textit{Concept Embedding Projection}, and
\textit{Concept Prediction Similarity} (see~\autoref{fig:teaser}). 
We support model comparison according to their produced word embeddings and classification predictions, i.e., both intrinsic and extrinsic evaluation methods.
The explanations are performed on diverse human-interpretable concepts related to bias mitigation and sentiment analysis tasks (e.g., \textit{gender-related stereotypes}, \textit{human qualities}).
The users can upload further concepts to the workspace to cover further analysis directions.
The modular composition of visual explanations supports such analysis extensions.

The comparison of adapter properties requires sufficient comparative visualization designs. 
As described by Gleicher~\cite{Gleicher2018ConsiderationsFV}, the design of comparative visualizations is not trivial since they typically combine the issues of representing individual objects as well as their relationships.
In order to design an appropriate solution, we rely on the comparative visualization guidelines~\cite{Gleicher2018ConsiderationsFV} and consider four task- and data-related aspects: (1) comparative elements, (2) challenges related to representing relationships between the comparative elements, (3) strategies to overcome the challenges, and (4) a sufficient design solution.
The design process constituted of several iterations in close collaboration with NLP researchers. 
In~\autoref{sec:design-rationale} we present some of the considered design alternatives; others are provided as supplementary material to this paper. 

We show the applicability of the workspace through case studies created collaboratively with NLP researchers.
In particular, we compare the properties of six adapters related to debiasing, sentiment classification, and named entity recognition tasks. 
\revision{We present new insights into model properties related to human-interpretable concepts and show that, for instance, context-0 (\textit{decontextualized}) embeddings of the adapter trained on the language debiasing task contain a bias where words become more similar to \textit{female-} than \textit{male pronouns}; however, the gender information is eliminated from the contextualized word representations.}

To summarize, the contribution of this paper is threefold. (1) We present requirements for a visual analytics system supporting fine-tuned LM comparison. (2) We introduce a workspace for model comparison and present design considerations for three types of comparative, visual explanation methods. (3) We present new insights into multiple adapter properties through expert case studies.

\begin{figure*}
    \centering
    \includegraphics[width=\linewidth,cfbox=light-gray 1pt 0pt]{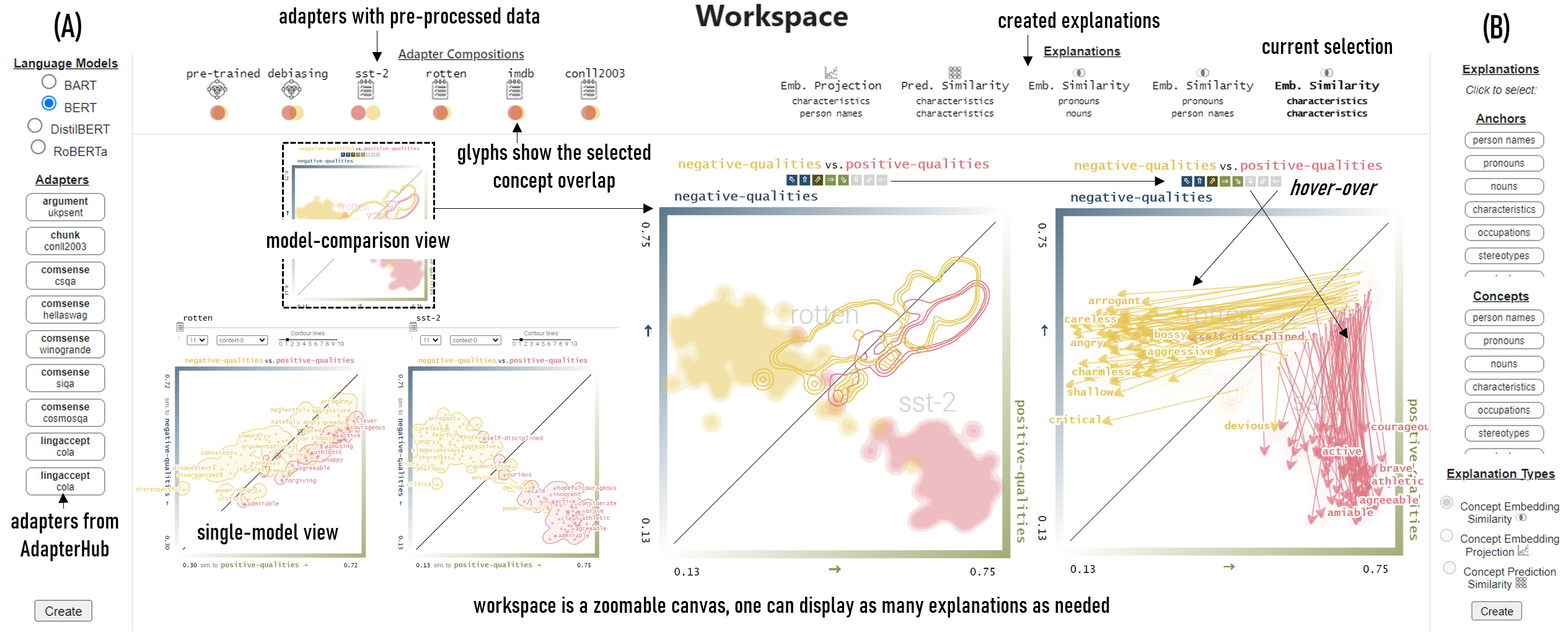}
    \vspace{-15pt}
  \caption{The workspace contains three views: \textbf{Adapter Composition View} (A), which lists adapters from AdapterHub repository, \textbf{Explanation Composition View} (B) for modular explanation generation, and \textbf{Visual Comparison View} (Workspace) for model comparison. 
  Here: contrary to the rotten-tomatoes model, the context-0 embeddings of the sst-2 sentiment classifier strongly encode the two polarities of \textit{human qualities}.
  }
    \label{fig:workspace}
    \vspace{-15pt}
\end{figure*}


\section{Background and Related Work}\label{sec:background}
In the following, we describe background information related to LM fine-tuning and related work to explanation methods.

\subsection{Language Model Fine-Tuning}
In this paper, we analyze transformers, which are multi-layer models that use attention mechanisms~\cite{10.5555/3295222.3295349}. 
In these models, each token of the input sequence is mapped to a high-dimensional vector (i.e., context-dependent embedding that encodes specific context properties). 
These embeddings are updated in each transformer's layer; thus, one can extract and analyze contextualized word embeddings layerwise (e.g., 12 layers for the BERT-base model).
It has been shown that these embeddings encode different language properties found in the training data\cite{rogers2020primer}.
LMs, including transformers, are commonly fine-tuned to capture language characteristics for specific domains or tasks. 
Domain-adaptive fine-tuning is an unsupervised fine-tuning approach based on a masked language modeling task on text from a specific domain~\cite{Han2019UnsupervisedDA}. 
Intermediate-task training is a model's fine-tuning on labeled data prior to task-specific fine-tuning~\cite{Phang2018SentenceEO}. 
Task-specific fine-tuning deals with adapting an LM to a particular output label distribution~\cite{howard-ruder-2018-universal}. 
The fine-tuning of LMs is effective yet time- and resource-consuming. 
Kirkpatrick et al.~\cite{doi:10.1073/pnas.1611835114} also showed that fine-tuning can lead to catastrophic forgetting of language characteristics acquired during the model's pre-training. 
To overcome these limitations, Houlsby et al.~\cite{houlsby2019parameter} introduced adapters. 
They are a lightweight alternative for model fine-tuning, only optimizing a small set of task-specific parameters learned and stored during the adaptation phase, thus, reducing both training time and storage space.
The AdapterHub framework~\cite{pfeiffer2020AdapterHub} has brought the advantage of a simple and efficient adapter composition and reuse -- one can upload their trained adapters to the AdapterHub or HuggingFace\footnote{\url{https://huggingface.co/}} repositories, and they are available in the framework for interested parties, supporting the open science practice. 
Adapters can be trained on masked language modeling as well as specific downstream tasks (e.g., sentiment classification). 
The trained adapters can be `attached' to the pre-trained model, leading to adapted model parameters.
The model with an attached task adapter can be used for the target task (e.g., sentiment classification).  
\revision{Adapters have been applied for tasks such as natural language generation~\cite{lin-etal-2020-exploring}, machine translation~\cite{philip-etal-2020-monolingual, kim-etal-2019-pivot}, domain adaptation~\cite{pham-etal-2020-study, glavas-etal-2021-training}, 
injection of external knowledge~\cite{lauscher-etal-2020-common}, and language debiasing~\cite{lauscher-etal-2021-sustainable-modular}.}

\subsection{Visual Embedding Explanation and Comparison}
\revision{With respect to explainability,  most relevant work has focused on visualizations that show \textbf{how} transformers work and \textbf{what} they learn. 
For example, visual analytics systems like NLIZE~\cite{liu2018nlize}, Seq2Seq-Vis~\cite{strobelt2018s}, BertViz~\cite{vig2019bertviz}, exBERT~\cite{hoover2019exbert}, SANVis~\cite{park2019sanvis}, and Attention Flows~\cite{derose2020attention} visualize the attention layer, i.e., to highlight tokens to which the model attends to in order to solve a task. 
Although widely used, attentions and their suitability for explanation purposes are being controversially discussed in related work (see, e.g.,~\cite{Jain2019AttentionIN}).
Other work has focused on visualizing word embeddings to show what LMs learn. The first such tools were designed for static embeddings, such as word2vec~\cite{mikolov2013efficient} and GloVe~\cite{pennington-etal-2014-glove}, and facilitated analogies \cite{liu2017visual} and tasks related to local word neighborhoods \cite{heimerl2018interactive}.  
Later, Berger~\cite{berger2020visually} explored correlations between embedding clusters in BERT~\cite{devlin2018bert}.} 
\revision{Recent tools focus on LM comparison tasks by visualizing multiple models simultaneously. 
For instance, Strobelt et al.~\cite{strobelt-etal-2021-lmdiff} present LMDiff -- a tool that visually compares LM probability distributions and suggests interesting text instances for the analysis. 
Heimerl et al.~\cite{heimerl2020embcomp} present embComb, which applies different metrics to measure differences in the local structure around embedding objects (e.g., tokens). 
Embedding Comparator by Boggust et al.~\cite{boggust2022embedding} is a system for embedding comparison through small multiples. 
It calculates and visualizes similarity scores for the embedded objects based on their local neighborhoods (i.e., shared nearest neighbors). 
Different from these two approaches, we provide explanations of pre-defined human-interpretable concepts, enabling testing more specific hypotheses related to embedding intersections. 
Sivaraman et al.\cite{sivaraman2022emblaze} present Emblaze, which uses an animated scatterplot and integrates visual augmentations to summarize changes in the analyzed embedding spaces. 
In contrast, we compare models by aligning the two spaces using juxtaposition, superposition, and explicit encoding techniques.
Our recent work called LMFingerprints~\cite{sevastjanova2022lmfingerprints} applies scoring techniques to examine properties encoded in embedding vectors and supports model as well as model layer comparison.}
Embedding comparison tasks are relevant for all types of data that get represented by embedding vectors.
For instance, Li et al.~\cite{li2018embeddingvis} present a visual analytics system for node embedding comparison (i.e., graph data), and Arendt et al.~\cite{arendt2020parallel} introduce a visualization technique called Parallel Embeddings for concept-oriented model comparison on image data, to name a few.


\section{Requirement Analysis}\label{sec:requirement-analysis}

Before designing the visual analytics workspace, we \revision{conducted a literature review related to LM comparison tasks (e.g.,~\cite{boggust2022embedding, sivaraman2022emblaze, heimerl2020embcomp}).} \revision{Furthermore, we conducted two semi-structured interviews in an online setting with two NLP researchers (co-authors of this paper) with expertise in language modeling tasks to discuss further common evaluation-related analysis aspects.} Our goal was to gather specific linguistically motivated analysis tasks and research challenges for \revision{the evaluation of} adapted LMs. 
In the following, we describe the gathered requirements through \textit{Models and Data} and \textit{Users and Tasks}~\cite{miksch2014matter}.

\subsection{Models and Data}

The NLP research focuses not only on developing and adapting new models with better performance but also on understanding the linguistic properties the models implicitly capture. 
Probing classifiers~\revision{\cite{jawahar-etal-2019-bert,linetal-2019-sesame,edmiston2020-BERT-morpho}} and adversarial testing~\revision{\cite{glockner2018,marvin-linzen-2018-targeted,richardson2019}} are the most common methods used in computational linguistics to understand such properties. 
The current research explores not only what the models learn but also when they fail and which limitations they have, such as different types of biases~\cite{garrido2021survey, mehrabi2021survey, blodgett-etal-2020-language}; as well as ways to mitigate those biases~\cite{ganin2015unsupervised, xie2017controllable, elazar2018adversarial, ravfogel2020null, ravfogel2022linear}. 
Visualizations are used to analyze the model latent spaces 
to gain insights into the degree of changes in embedding vectors~\cite{ethayarajh2019contextual, sevastjanova2021}, properties encoded in embedding vectors~\cite{sevastjanova2022lmfingerprints}, and word neighborhood changes~\cite{heimerl2020embcomp, boggust2022embedding, sivaraman2022emblaze}. 
\revision{Especially, the comparison of embedding local neighborhoods is one of the critical tasks for many users of LMs~\cite{boggust2022embedding, sivaraman2022emblaze}. For such comparisons, one first needs to select words for the analysis. 
Boggust et al.~\cite{boggust2022embedding} write that this is commonly done either in a data- or model-driven way, for instance, by exploring specific domain-related words or challenging words for the analyzed model.
During the interviews, the NLP researchers agreed with this statement and emphasized that evaluation methods related to model limitations often explore specific, pre-defined human-interpretable concepts such as \textit{gender-related stereotypes}.
When analyzing such human-interpretable concepts, people commonly analyze contextualized word embeddings. 
For some methods (e.g., Word Embedding Association Tests~\cite{doi:10.1126/science.aal4230}), researchers compute word-level vectors without an explicit context~\cite{lauscher-etal-2021-sustainable-modular, 10.1162/coli_a_00391}. 
In particular, for BERT, \revision{one} can append the sequence start and the separator token before and after the word, respectively (e.g., [CLS] word [SEP]) and extract embeddings with context size zero~\cite{zhao2020quantifying} (also known as \textit{decontextualized} embeddings~\cite{bommasani-etal-2020-interpreting}). In the following, we call them context-0 embeddings. 
Our experts also emphasized the need to `connect' the embedding space with the model's behavior to inspect whether specific embedding vectors influence the model's predictions on downstream tasks.}

\subsection{Users and Tasks}

With this work, we aim to support developers and researchers who adapt and evaluate LMs to perform their analysis more easily \revision{by focusing on the analysis of diverse human-interpretable concepts}. 
To do that, we gathered task-related requirements.
NLP researchers' work is related to comparison (i.e., baseline) tasks. 
\revision{In particular, their analysis typically involves \textbf{(T0)} a comparison of multiple LMs with different architectures or fine-tuning settings as well as multiple model layers.}
Second, they typically analyze specific human-interpretable concepts and try to
\textbf{(T1)} partition the representation (e.g., embedding) space according to these concepts.
Third, they try to \textbf{(T2)} understand interactions between specific concepts, e.g., to what extent these concepts are represented similarly in the representation (e.g., embedding) space. 
They aim to \textbf{(T3)} detect `unexpected' associations, e.g., positive sentiment words that tend to trigger the negative sentiment because, e.g., they are negated. 
\revision{And finally, their goal is to \textbf{(T4)} connect the representation space with the actual behavior of the model, e.g., to understand whether concepts are separated in the representation space yet do not affect the behavior of the model.}


\section{Visual Analytics Workspace: Data Processing}
In this section, we present our visual analytics workspace and its three main components: \textbf{Adapter Composition View} (in~\autoref{fig:workspace} A), \textbf{Explanation Composition View} (in~\autoref{fig:workspace} B), and \textbf{Visual Comparison View} (in~\autoref{fig:workspace} Workspace) for model and layer comparison. Before introducing the workspace design in \autoref{sec:design-rationale}, we describe the data processing.

\subsection{Data Modeling}\label{sec:data-modeling}
Motivated by the gathered requirements, we first build the data model. 
Since human-interpretable concept analysis plays a crucial role in NLP research, we start by modeling such concepts.  
By default, we work with concepts that are commonly used in research related to bias mitigation\footnote{\url{https://github.com/cisnlp/bias-in-nlp}} and sentiment analysis. 
The users can upload further concepts as \revision{.json} files in the interface.
One concept is represented by two word lists, each having a specific polarity. 
For instance, a concept called \textit{person names} consists of two word lists -- \textit{male person names} and \textit{female person names}, respectively.
\revision{We provide the following concepts: \textit{male/female person names}, \textit{male/female pronouns}, \textit{male/female-related nouns}, \textit{male/female-related stereotypes}, \textit{positive/negative human qualities}, \textit{high/low-GDP countries}, and words related to \textit{weak/strong}, \textit{family/career}, \textit{science/arts}, \textit{intelligence/appearance}}.

We first model each word in a concept through a list of sentences in which the word is used. 
For this purpose we use the Yelp dataset~\cite{zhangCharacterlevelConvolutionalNetworks2015}; the user can also upload other datasets and use them for explanations. 
The associated sentences are used for two purposes. 
First, we use them as an input to the (adapted) LM to extract the word's contextualized word embeddings.
The embeddings are extracted layerwise (i.e., layer 1-12 for BERT-base) and get aggregated~\cite{bommasani-etal-2020-interpreting} for each unique word (e.g., one average embedding from all occurrences of the word \textit{Germany} per layer).
Second, we use these sentences as input for task adapters for prediction making. 
Furthermore, we extract the word's context-0 embedding by using the model's special tokens and the word itself as the input to the model (i.e., [CLS] word [SEP]).
For words that do not occur in the vocabulary, we average their sub-token embeddings.

\subsection{Adapter Composition and Explanation Composition}\label{sec:adapter-composition}
We load adapters from AdapterHub repository and list them in the \textbf{Adapter Composition View}. 
The user can select an adapter for the analysis by clicking on the particular icon.
Currently, we have pre-processed the data for six models: the \textbf{pre-trained} BERT (BERT-base-uncased), the \textbf{debiasing} BERT~\cite{lauscher-etal-2021-sustainable-modular}, and four task adapters for BERT (sentiment classifiers \textbf{sst-2}, \textbf{rotten-tomates}~\cite{poth-etal-2021-pre}, and \textbf{imdb}~\cite{poth-etal-2021-pre}, and the named entity recognizer \textbf{conll2003}).
For a new adapter selection, the data is first pre-processed and stored in the database.

The user defines which explanation methods to use for their analysis in the \textbf{Explanation Composition View}. 
The explanations are constructed from available concepts and three visualization types. 
The visualizations include \textit{Concept Embedding Similarity}, \textit{Concept Embedding Projection}, and \textit{Concept Prediction Similarity}. 
The Concept Embedding Similarity requires an input of two concepts: one is used as an anchor in the visualization and the other is explained through the cosine similarity to the anchor. 
The Concept Embedding Projection requires an input of one or two concepts (to analyze a single concept or the relation between two (un)related concepts). 
The user can choose between multiple projection techniques: 
Principal Component Analysis (PCA)~\cite{jolliffe2016principal}, Multidimensional Scaling (MDS)~\cite{kruskal1964multidimensional}, t-Distributed Stochastic Neighbor Embedding (t-SNE)~\cite{van2008visualizing}, and Uniform Manifold Approximation and Projection (UMAP)~\cite{mcinnes2018umap}. 
The Concept Prediction Similarity can be applied only on adapters with prediction heads (e.g., sentiment classifier). 
The explanation requires an input of one concept; the class labels are used as anchors in the visualization. 

The pre-computed adapters, as well as created explanations, are displayed on top of the \textbf{Visual Comparison View}, represented through an icon and adapter's or explanation's name. 
The user first selects an explanation type, then an adapter that they would like to analyze. 
To guide the users toward interesting adapters for the analysis, we display a glyph underneath the adapter's icon. 
The glyph shows the overlap between the two concept word lists for the selected explanation.
The overlap is determined using a similar algorithm to the class consistency~\cite{sips2009selecting} that is commonly used to select good scatterplot views for high-dimensional data.
An example of these glyphs is shown in~\autoref{fig:workspace}.
The explanation visualization is displayed in the \textbf{Visual Comparison View} on a zoomable canvas\revision{; hence, one can display as many explanations on the canvas as needed}. 
A draggable placeholder icon \includegraphics[height=10pt]{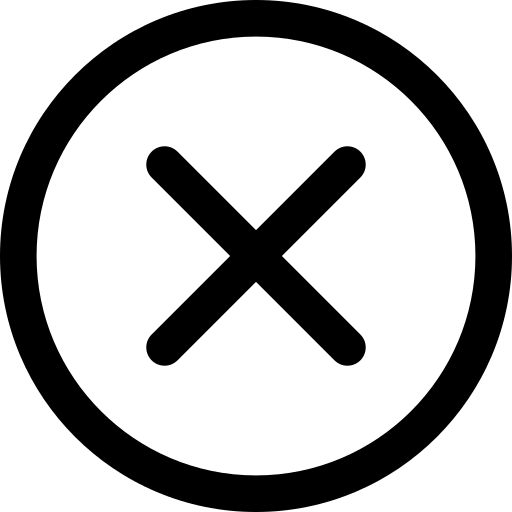} marks the position where the next selected adapter visualization will be displayed on the screen.

\section{Visual Analytics Workspace: Design Rationale}\label{sec:design-rationale}

\begin{figure*}
    \centering
    \includegraphics[width=0.95\textwidth]{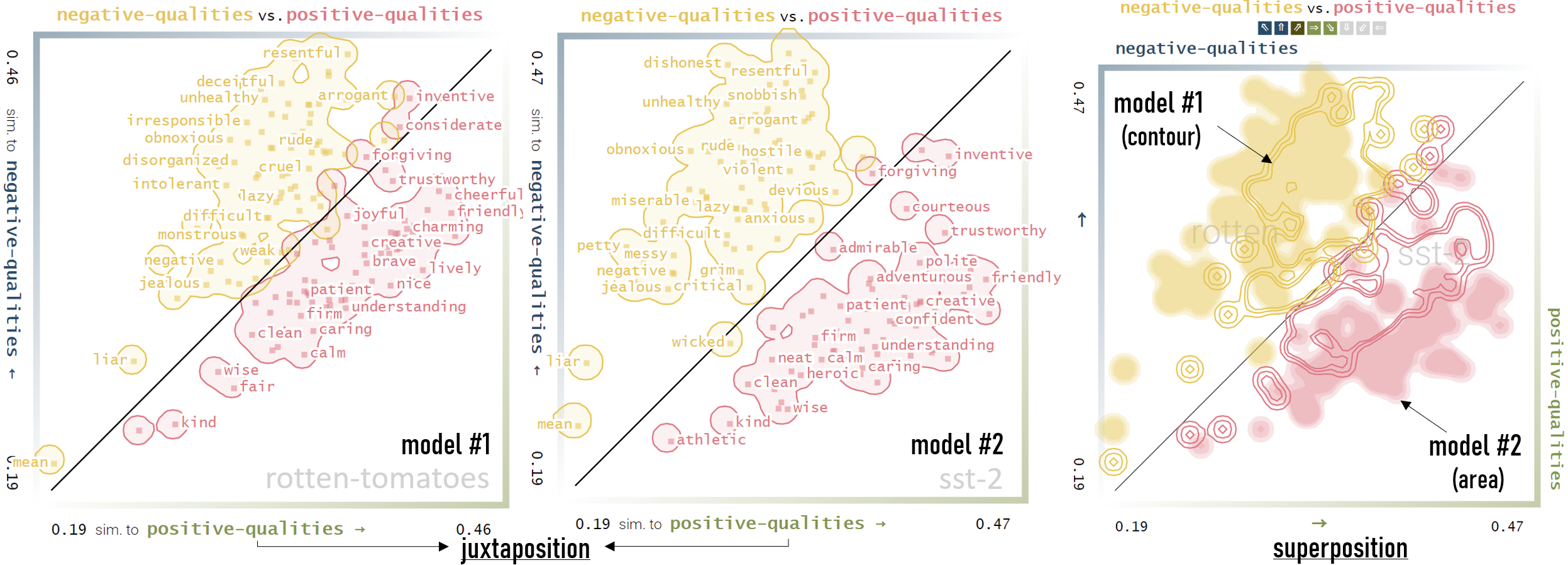}
     \vspace{-8pt}
    \caption{
    We provide two types of model comparison designs for analyzing concept embedding \textbf{similarity}, i.e., \textbf{juxtapositon} where two models are displayed next to each other and \textbf{superposition}, where two models are displayed in one visualization.
    Here: the contextualized word embeddings extracted from layer 11 for the rotten-tomatoes and sst-2 sentiment classifiers differentiate between \textit{positive-} and \textit{negative human qualities}. The rotten-tomatoes model requires context to separate the two polarities since the separation is stronger than for context-0 embeddings (see~\autoref{fig:workspace}).
    }
    \label{fig:positions-superposition}
    \vspace{-15pt}
\end{figure*}

In the following, we describe the design rationale and the visual encoding for the designed explanation visualizations. 
Our workspace supports the exploration of a single model and the comparison of two models or two model layers \textbf{(T0)}. We apply diverse explanation methods (i.e., the similarity in the high-dimensional space, embedding projection, and explanation details) to detect and avoid potential artifacts generated by a single approach (e.g., projection artifacts).
The design of the comparison visualizations was motivated by the design guidelines by Gleicher~\cite{Gleicher2018ConsiderationsFV} that consider the comparative elements, challenges that may occur, strategies to overcome the challenges, and the design solutions.

\paragraph{Global Visual Encoding} 
In all visualizations, we use the visual mark called point~\cite{carpendale2003considering} (i.e., rectangle) to represent words. Hidden word labels are displayed by hovering over a word's rectangle.
We use positional encoding~\cite{carpendale2003considering} to partition the embedding space \textbf{(T1)}, detect concept intersections \textbf{(T2)}, and locate ‘unexpected’ associations \textbf{(T3)}. The position is used to show the similarity between words according to underlying features such as different types of word embedding vectors or prediction labels. 
We group words belonging to the same concept through an additional visual mark, i.e., area/contour.
The contours are implemented using the d3-contour library\footnote{\url{https://github.com/d3/d3-contour}} based on a two-dimensional kernel density estimation on the point clouds. 
The user can specify how many contour lines to display in the visualization by moving a slider. 
To support memorization and ease the readability, we use a global \begin{wrapfigure}[7]{r}{0.15\textwidth}  
    \centering
    \vspace{-14pt}
    \includegraphics[width=\linewidth]{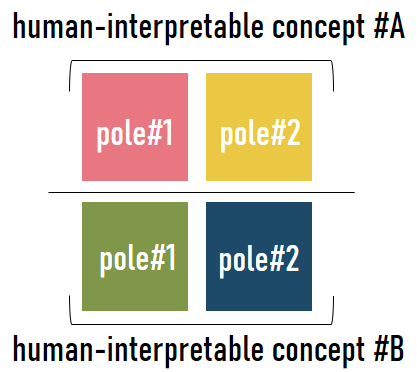}
\end{wrapfigure}color encoding~\cite{carpendale2003considering} for concepts.
In particular, we use two diverging color pairs.
One color pair represents the two word lists of a concept. 
The selection of the color pairs was not trivial since the colors had two objectives: the separability between two concepts and the separability between two word lists of one concept. The final decision was made as follows: we selected two warm colors (i.e., \textcolor{pink}{\textbf{pink}} and \textcolor{yellow}{\textbf{yellow}}) representing one concept and two cold colors (i.e., \textcolor{green}{\textbf{green}} and \textcolor{blue}{\textbf{blue}}) representing the other, as shown in the side figure. Further color alternatives are included in the supplementary material.

\paragraph{Visual Encoding for Single Model Visualizations}
By default, we display as many details as possible in the single visualizations but avoid label overplotting. 
An algorithm measures whether displaying a label would lead to overlap. 
\revision{The algorithm iterates through words in both word lists of a concept and measures the bounding box of each text element that gets added to the visualization. If the new element creates an overlap, it is hidden in the visualization.}

\paragraph{Visual Encoding for Model Comparison Visualizations}
For effective model comparison, we use both the juxtaposition design (see~\cite{Gleicher2018ConsiderationsFV}) and either the superposition for visualizations that have a positional anchor or explicit encoding for visualizations that lack the positional anchor (e.g., projection techniques).
By default, we show the summary~\cite{Gleicher2018ConsiderationsFV} of the two models to avoid datapoint overplotting. The summaries are created using the contour library; the source model is represented through its contour in the 2D space, and the target model is represented through its filled-out area.
We use the scan sequentially~\cite{Gleicher2018ConsiderationsFV} strategy to show exact word positions. The filter icons are explained in~\autoref{sec:emb-sim}.

\subsection{Concept Embedding Similarity}\label{sec:emb-sim} This explanation displays the cosine similarity between two concepts enabling to partition the embedding space \textbf{(T1)}, detect concept intersections \textbf{(T2)}, as well as locate `unexpected' associations \textbf{(T3)}.
In this representation, one concept is used as an anchor for explanation purposes. 
The other concept can be the same as the anchor (e.g., \textit{human qualities} used twice in~\autoref{fig:positions-superposition}) or it may differ from the anchor (e.g., \textit{person names} as a concept and \textit{pronouns} as an anchor in~\autoref{fig:debiasing}).
We measure the average cosine similarity between a word in the concept to words in each pole of the selected anchor.
It helps to analyze different biases in the data, for instance, whether, e.g., \textit{female pronouns} are more similar to specific \textit{stereotype} words than \textit{male pronouns}. 

\textbf{(1) Single Model Explanation --} The two anchor word lists represent the two axes in the scatterplot visualization (e.g., \textit{negative qualities} represent y-axis and \textit{positive qualities} represent x-axis in~\autoref{fig:positions-superposition}). 
The average similarity values between a word in the concept to the anchors are used as coordinates in the 2D visualization. 
A word's (e.g., \textit{cheerful} in~\autoref{fig:positions-superposition}) average similarity to the first anchor word list (e.g., \textit{negative qualities}) specifies the word's y-position and the average similarity to the second anchor word list (e.g., \textit{positive qualities}) specifies the word's x-position.
To support the readability, we add a diagonal line to the visualization as a point of reference.
If a word is more similar to the first word list, then it will be located on the left-hand-side of the diagonal; if a word is more similar to the second word list, then it will be located on the right-hand-side of the diagonal.
Words that are equally similar to both word lists are located on the diagonal.
By default, we display all words in the concept word lists as rectangles and show non-overlapping labels.
Since \revision{most of the} word lists consist of ca. 100 words, the visualization has overplotting issues that limit the analysis of concept intersections.
To overcome these issues, we add a contour line around each pole.
We use the d3-contours library and specify the bandwidth parameter to 5, which leads to larger areas for more dense regions; however, single outlier data points are enclosed in separate, smaller areas, enabling the detection of `unexpected' associations \textbf{(T3)}.
The area is colored in the particular concept's color with a decreased opacity.

\begin{figure*}[h]
  \begin{center}
   \begin{subfigure}[b]{0.92\textwidth}
        \includegraphics[width=\textwidth]{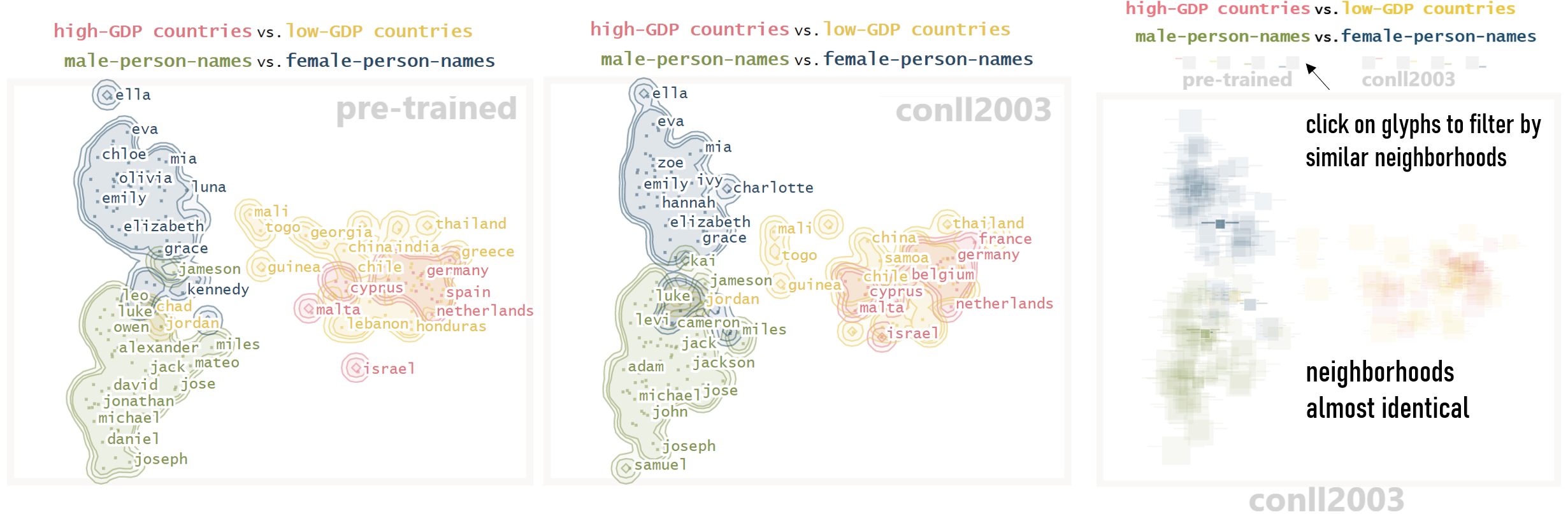}
        \vspace{-15pt}
        \caption{In layer 11, the PCA projection generates almost identical 2D spaces for contextualized embeddings extracted from pre-trained BERT and conll2003 named entity recognizer (see the low opacity of word rectangles in the plot on the right hand side). In both models, the \textit{person names} get separated by gender.}
        \label{fig:layer1}
    \end{subfigure}
    \begin{subfigure}[b]{0.92\textwidth}
        \includegraphics[width=\textwidth]{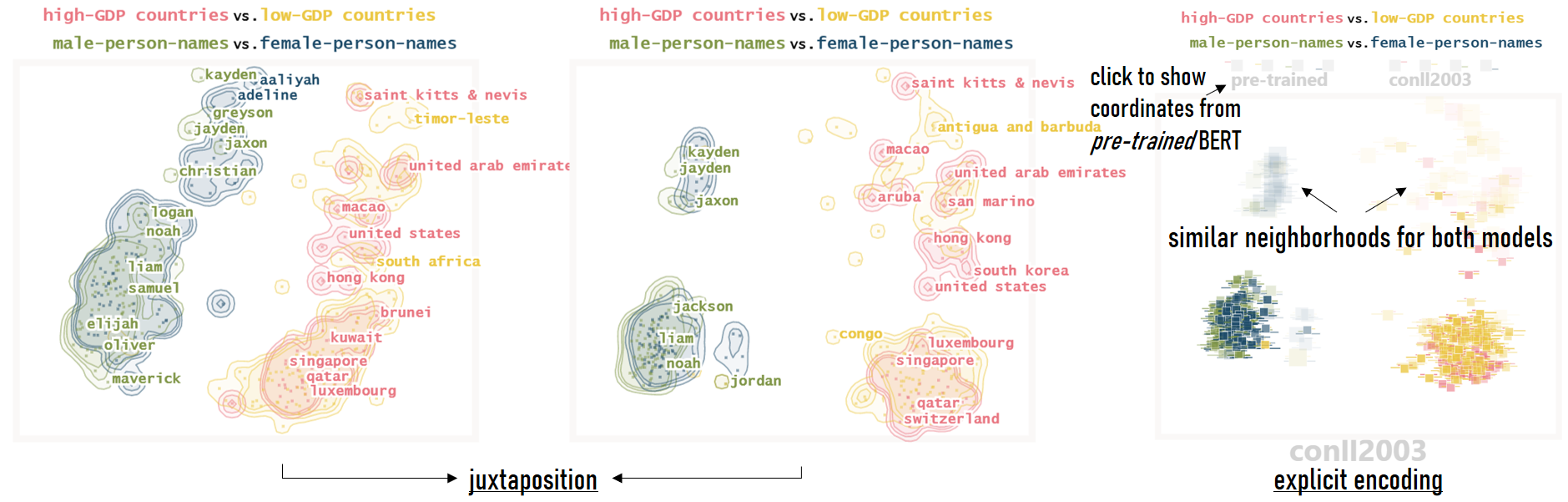}
        \vspace{-15pt}
        \caption{In layer 11, the PCA projection of context-0 embeddings from conll2003 named entity recognizer produces four distinct clusters. Two clusters (with low opacity) have similar neighborhoods in both models. These are rare person names (e.g., Nevaeh) and long country names (e.g., Trinidad and Tobago). \textit{Person names} do not encode gender.}
        \label{fig:layer11}
    \end{subfigure}
    
    \vspace{-6pt}
    \caption{ 
    We provide two different types of model comparison designs for analyzing concept embedding \textbf{projections}, i.e., \textbf{juxtapositon} where two models are displayed next to each other and \textbf{explicit encoding} that summarizes embedding changes through word neighborhood overlaps. }
     \label{fig:projection}
          \end{center}
 \vspace{-25pt}
\end{figure*}

\textbf{(2) Model Comparison Explanation --} 
As mentioned in~\autoref{sec:requirement-analysis}, the overall goal of NLP researchers is to compare models or layers with respect to concept distributions \textbf{(T0)}.
The design of comparison visualizations is not trivial, as described by Gleicher~\cite{Gleicher2018ConsiderationsFV}.
Thus, in order to consider all relevant aspects, we follow his design guidelines. 

The comparison visualization for Concept Embedding Similarity has to display two models or layers simultaneously, each showing the distribution of concept words with respect to selected anchors. 
Two types of challenges may arise when designing for this objective: (1) the concepts, as well as models, may overlap, and (2) word similarity changes may produce patterns that are difficult to outline all at once.
Before we describe the strategies to overcome these challenges, we name our design considerations.
Gleicher~\cite{Gleicher2018ConsiderationsFV} names three design alternatives for comparison visualizations: juxtaposition, superposition, and explicit encoding.
In our workspace, each explanation can be explored in a juxtaposition design (shown in~\autoref{fig:positions-superposition} left) since single model visualizations are always displayed next to each other on the screen.
This representation has limitations, though. 
Since we use all the available 2D space for a single model to reduce word overlaps, the visualizations of the compared models often have different scales. 
Thus, the detailed model and concept overlap analysis is restricted.
Therefore, instead of using juxtaposition, we place two models in the same representation using the superposition design (shown in~\autoref{fig:positions-superposition}, right).
The superposition is a valid alternative since the Concept Embedding Similarity visualization has anchors (which is not the case for projection techniques, as described in the following).

In the comparison visualization, we display the cosine similarity values between concept words and anchors for two models simultaneously \textbf{(T0)}.
We follow the comparative visualization guidelines and apply two strategies that enable the analysis of overlapping concepts, models, and word similarity patterns.
First, we provide a summary of the two models.
We, therefore, display only the contours of their word positions; more details (e.g., word exact positions) are displayed on demand.
During the design process, we created several alternative representations \revision{to visually separate the two models}. 
Each designed alternative was discussed with a group of visual analytics experts to critically assess the representation's advantages and limitations.
In particular, we created representations that showed \revision{two} types of the density of the visualized words, i.e., discrete as well as continuous. 
The discrete representation displayed the density regions through triangles arranged on a grid layout, whereby each model was represented with triangles of different sizes and opacity (smaller rectangles with higher opacity for the target \begin{wrapfigure}[10]{r}{0.235\textwidth}  
    \centering
    \vspace{-11pt}
    \includegraphics[width=\linewidth]{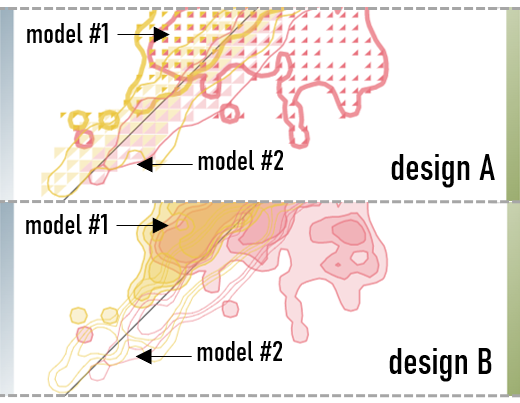}
\end{wrapfigure}model, \revision{see design A in the side figure}).
The continuous representation summarized the models through their contours \revision{(see design B in the side figure)}.
After several discussions, the latter was selected as the final design due to its visual smoothness and limited clutter.
The final design is as follows: the first (i.e., source) model is displayed only through contour borders. 
Since the words themselves are not visible, we use multiple contour lines to highlight the density of the word-occurrence regions. 
The second (i.e., target) model is displayed through a filled-out area of the contour regions with transparency. 
In addition to the model summarization, we apply the scan sequentially strategy to enable the analysis of word similarity changes. 
For this purpose, we implemented filter buttons that can be used to highlight words that have common properties with respect to their positional changes (i.e., their position in the source model compared to their position in the target model in the 2D space). 
In particular, we measure the angle between the word's position in the source and the target model.
By hovering over one of the filter buttons \includegraphics[height=10pt]{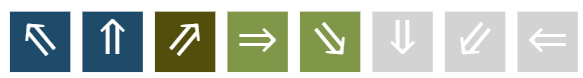}, words with similar positional changes are highlighted in the visualization.
\revision{The buttons themselves are colored according to the anchor to which words in the target model become more similar in comparison to the source model.}
An example of the word filtering is shown in~\autoref{fig:workspace}.

\vspace{-6pt}
\subsection{Concept Embedding Projection} The second explanation method displays the words in a 2D visualization, whereby the 2D positions are obtained using a projection technique such as PCA on the embedding vectors. 
This explanation visually partitions the representation space \textbf{(T1)} and supports the analysis of concept intersections \textbf{(T2)}.
Since in the Concept Embedding Similarity explanation we compute the similarity on high-dimensional vectors, this representation shows the similarity from a different modeling perspective.

\textbf{(1) Single Model Explanation --}
The explanation displays words within one or two concepts, depending on whether the user wants to analyze one concept or the overlap of two (un)related concepts.
Like in every visualization, we display words as rectangles and, by default, show labels for words that do not overlap.
To support the readability of dense regions, we designed and discussed several design alternatives.
First, we displayed words using a scatterplot technique, which is common for displaying projection data (design A in the side figure).
Since the goal of the visualization is to clearly show concept intersections \textbf{(T2)}, however, words in the projection often overlap, this representation was not feasible.
Second, we applied a kernel density estimation algorithm on the projected words to estimate and visualize the densest regions in the 2D space.   
We \begin{wrapfigure}[9]{l}{0.29\textwidth}  
    \centering
    \vspace{-12pt}
    \includegraphics[width=\linewidth]{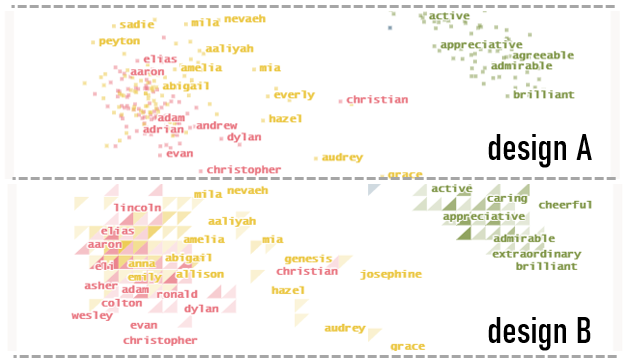}
\end{wrapfigure}first represented the density through triangles displayed in a grid layout, whereas the density value was mapped to the triangles' opacity (design B in the side figure).
Similar to the simple scatterplot, it was difficult to detect concept intersections easily.
Thus, in the final design, we use multiple contours showing the estimated density of the different regions (\autoref{fig:projection}). 
It allows detecting not only the densest regions but also words with unexpected associations \textbf{(T3)} (i.e., outliers).

\textbf{(2) Model Comparison Explanation --} 
Our goal is to display intersections and positional changes of one or two concept word lists. 
The challenge of this representation is grounded in the artifacts of the applied projection techniques.
In particular, since we rely on projection techniques to compute word coordinates, the visualization lacks an interpretable point of reference; projection techniques typically come with artifacts such as rotation or flipping of the representation space, making the comparison of two spaces difficult. 
Like in all other visualizations, the user can explore model differences in a juxtaposition design since the single model explanations are always placed next to each other on the screen (as shown in~\autoref{fig:layer11}, left).
The juxtaposition has limitations, though.
If the compared models produce different embedding spaces (which is the case for most of the model and layer comparisons), they produce 2D spaces that are difficult to align. 
The insufficiency of the superposition design is depicted in the side figure. There, we represent a word's positional changes through lines, \begin{wrapfigure}[8]{r}{0.15\textwidth}  
    \centering
    \vspace{-12pt}
    \includegraphics[width=\linewidth]{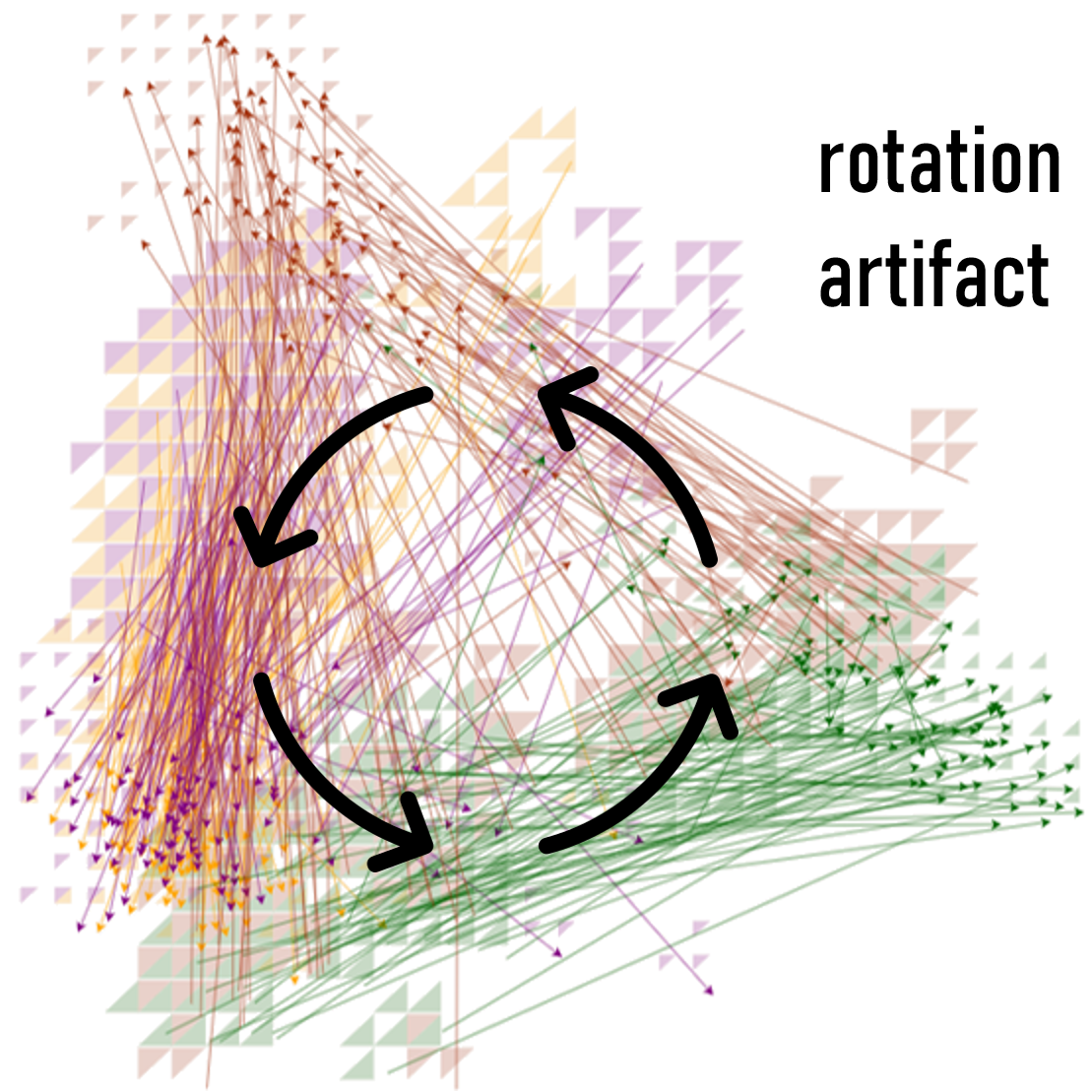}
\end{wrapfigure}whereas a line connects the word's position in the source model with the position in the target model.
Due to rotation artifacts, the comparison of word changes is restricted even if the changes are minor.
Thus, for projection comparison purposes, we apply the third design alternative, i.e., the explicit encoding design (as shown in~\autoref{fig:layer11}, right).

\begin{figure}
    \centering
    \includegraphics[width=\linewidth]{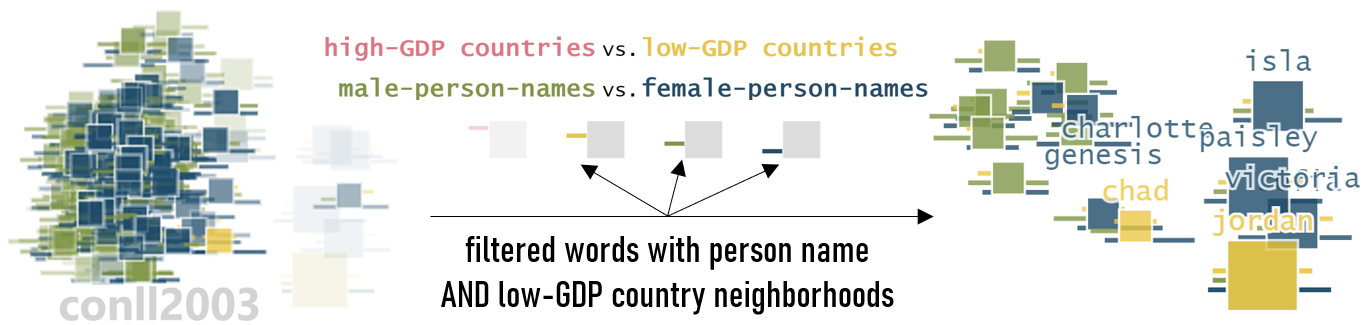}
     \vspace{-14pt}
    \caption{Words with similar neighborhoods can be filtered by selecting particular glyphs. In conll2003 named entity recognizer, country names \textit{Jordan} and \textit{Chad} are more similar to \textit{person names} than \textit{countries}. }
    \label{fig:names-filtered}
    \vspace{-18pt}
\end{figure}

For the explicit encoding, we first define relationships to encode in the visualization~\cite{Gleicher2018ConsiderationsFV}, i.e., we explain the projection changes through word nearest neighbors in the 2D space.
In particular, after computing the projection's coordinates, we compute ten nearest neighbors for each word and store them as attributes in the data structure.
When the user explores two models according to their embedding projections, we visually explain the neighborhood overlaps. 
This, according to design guidelines~\cite{Gleicher2018ConsiderationsFV}, is an example of the summarize strategy.
\revision{Unlike} the Concept Embedding Similarity visualization, we display only a single word's instance in the visualization. 
Its 2D coordinates, by default, are coordinates from the source model. 
The user can change it by clicking on the model's name in the visualization (shown in~\autoref{fig:layer11}, right).
The neighborhood changes are displayed as follows. 
For each word, we measure the neighborhood overlap (the number of equal neighbors in the source and target model) and map it to the size of the word's rectangle representation. 
The higher the overlap, the larger the rectangle and the lower the opacity.
Moreover, we add horizontal \begin{wrapfigure}[2]{r}{0.08\textwidth}  
    \centering
    \vspace{-13pt}
    \includegraphics[width=\linewidth]{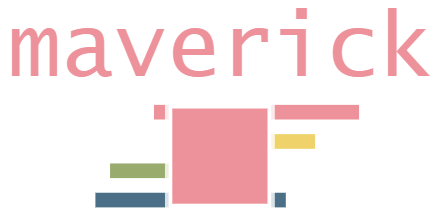}
\end{wrapfigure}lines to the rectangle, each showing the nearest neighbors from the particular concept's pole. 
As shown in the side figure, in the pre-trained BERT the person-name \textit{Maverick} is more similar to \textit{countries} (\textcolor{blue}{\textbf{blue}} and \textcolor{green}{\textbf{green}} lines on the left-hand-side) than \textit{person names}; in the conll2003 named entity recognizer, this word becomes more similar to \textit{person names} (\textcolor{yellow}{\textbf{yellow}} and \textcolor{pink}{\textbf{pink}} lines on the right-hand-side of the rectangle). 
An example of two models with similar word neighborhoods is shown in~\autoref{fig:layer1} and with different word neighborhoods -- in~\autoref{fig:layer11}. 
If the word neighborhoods change, then rectangles are smaller with a higher opacity, as shown in~\autoref{fig:layer11}. 
In addition to the summarize strategy, we support the scan sequentially strategy to enable the analysis of word neighborhood changes.
The users can filter words based on their neighborhoods by clicking on the glyph representations displayed on top of the visualization.
The filtered words are highlighted; the rest are faded out (shown in~\autoref{fig:names-filtered}).
On mouse over a word, its nearest neighbors in the source model are highlighted; on click, the nearest neighbors in the target model are highlighted, enabling a simple neighborhood comparison.

\subsection{Concept Prediction Similarity}\label{sec:prediction}
The third visualization can be used on adapters that have been trained on two-class classification tasks.
It explains the prediction similarity of two models that are trained on the same task, e.g., whether two sentiment classifiers produce similar prediction outcomes, and connects the representation space and the model's behavior \textbf{(T4)}.
For this task, the user has to select one concept; the model then predicts class labels for the words' assigned sentences.

\textbf{(1) Single Model Explanation --}
To provide an overview of prediction similarity, we aggregate the label information for all sentences in which the word is used in the corpus and use the average prediction to determine the word's x-coordinate in the visualization.
In particular, we divide the number of sentences having the first prediction label (e.g., \textcolor{blue}{\textbf{NEGATIVE}} sentiment) by the total number of sentences for the particular word; the more predictions with the first class label -- the closer the point is to the beginning of the x-axis. 
If the predictions are equal for both class labels, the word is placed in the middle of the x-axis.
The y-coordinate is determined by the word's position in the particular word list.
The words themselves are displayed as rectangles.

\begin{figure}[b]
    \vspace{-14pt}
    \centering
    \includegraphics[width=\linewidth]{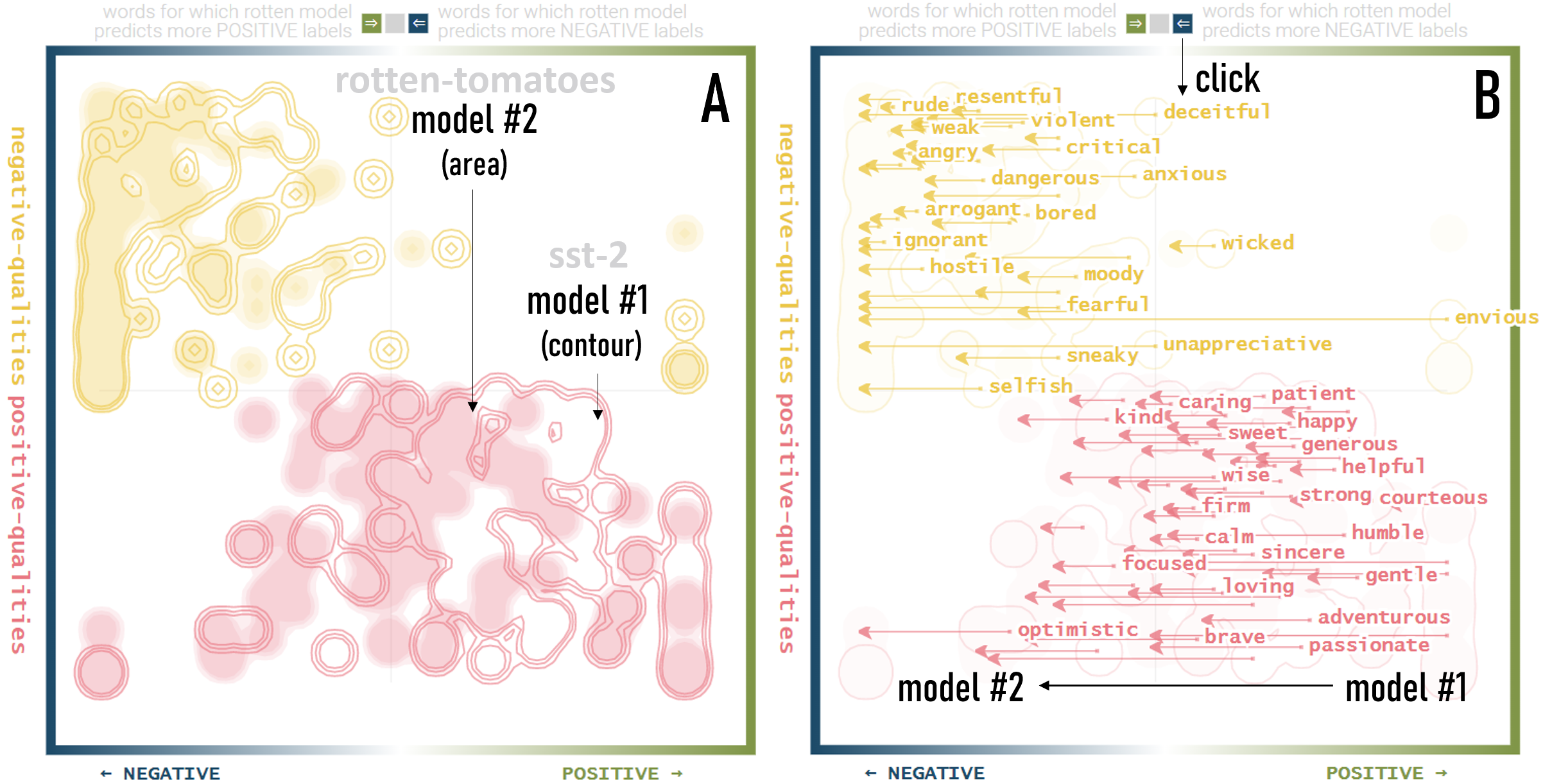}
    \vspace{-12pt}
    \caption{\revision{Concept Prediction Similarity shows two sentiment classifiers (see A). Compared to the sst-2 model (contour borders), the rotten-tomatoes model (filled areas) classifies sentences with occurrences of \textit{positive} and \textit{negative human qualities} more often as \textcolor{blue}{\textbf{NEGATIVE}} (B).}
}
    \label{fig:comparison-predictions}
\end{figure}

\textbf{(2) Model Comparison Explanation --} 
In the comparison visualization, our goal is to show the prediction differences between two models \textbf{(T0)}.
Since in this visualization we have clear anchors (the prediction labels), we can apply a similar design approach as for the Concept Embedding Similarity plot. 
In particular, we use both juxtaposition as well as superposition designs.
In the superposition design, both models are represented in the same visualization, as shown in~\autoref{fig:comparison-predictions}.
We stick to the same design as for the Concept Embedding Similarity plot and first summarize the model predictions through contours.
The source model is represented through the contour's borders; the target model's contours are filled out with a decreased opacity.
The user can click on the filtering icons displayed on top of the visualization; the prediction changes are highlighted accordingly, supporting the scan sequentially strategy.

\subsection{Explanation Details}
When explaining model changes, researchers usually try to find the reasons for particular patterns in the data.
\revision{Thus,} we designed three visualizations to explain patterns in the comparison visualizations.

\textbf{Context Concordance View --}
The patterns in the Concept Embedding Similarity visualization can be influenced by the word contexts (sentences) from which the contextualized word embeddings are extracted.
Thus, for this visualization, we added a \textit{Context Concordance View} that lists all sentences in which a word is used in the corpus (shown in~\autoref{fig:teaser}, right).
The view is displayed when clicking on the particular word in the Concept Embedding Similarity visualization. There, the selected word is highlighted for a better comparison.

\textbf{Projection Artifact View --}
We propose a dense pixel visualization to explore the latent space and reveal semantically similar embeddings.
The pixel visualization is inspired by Shin et al.~\cite{shin2018interpreting} stripe-based visualization of word embeddings.
The primary goal is to create a compact visual summary of the embeddings with all dimensions without using dimensionality reduction methods (e.g., PCA).
The pixel visualization displays each embedding as a vertical pixel bar, a grid-shaped column where each colored pixel (rectangle) is an embedding feature value.
Herefore, we normalize the embeddings to the unit length and color the pixels according to a diverging color scheme. 
Then we place the pixel bars next to each other on the x-axis, producing a dense pixel visualization.
The y-axis displays the 768 embedding dimensions, and the rows are ordered by the median of the visualized embedding dimensions to highlight block and band patterns~\cite{behrisch2016matrix}.
The x-axis can be reordered by linking and brushing in the single model explanations to interactively create clusters to highlight and display as a block of embeddings.
Alternatively, the embeddings can be clustered using HDBSCAN~\cite{campello2013density} using cosine similarity to detect clusters of similar embeddings.
We can explore clusters in latent space through clustering without relying on dimensionality reduction methods, which typically produce some artifacts.
Overall, comparing the colored pixel bars enables us to perceive pairwise similarities between the embeddings and generate new insights into the latent space, such as identifying groups of similar embeddings, meaningful embedding dimensions, or outliers. 

\textbf{Prediction View --} 
To explore the exact prediction differences in the Concept Prediction Similarity comparison visualization, we display the predicted labels for all sentences assigned to a word in the \textit{Prediction View} (shown in~\autoref{fig:teaser}, right). 
The view is displayed when selecting a word in the Concept Prediction Similarity visualization.

\begin{figure*}
    \centering
    \includegraphics[width=\linewidth]{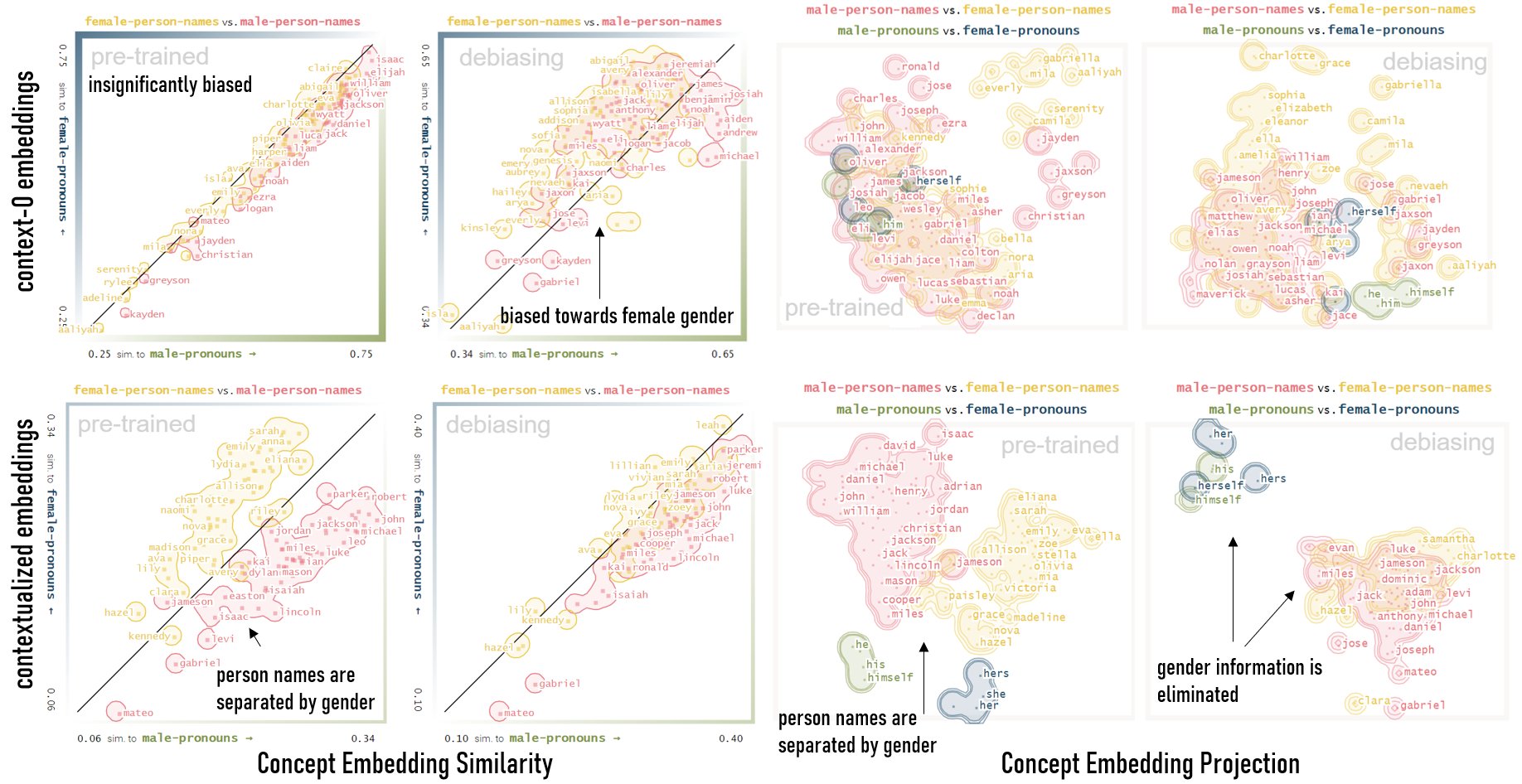}
    \vspace{-18pt}
    \caption{\revision{Context-0 embeddings are used for evaluation purposes in Word Embedding Association Tests~\cite{10.1162/coli_a_00391,lauscher-etal-2021-sustainable-modular}. Their produced spaces differ from the contextualized ones, though. Although context-0 embeddings suggest that the debiasing adapter by~\cite{lauscher-etal-2021-sustainable-modular} inverts the gender bias of the pre-trained BERT, the PCA projection on contextualized embeddings shows that the adapter successfully eliminates the gender information.}}
    \label{fig:debiasing}
    \vspace{-15pt}
\end{figure*}

\section{Evaluation}
\revision{We conducted expert case studies~\cite{sedlmair_design_2012} with the experts from the requirement analysis (see~\autoref{sec:requirement-analysis}) to assess initial feedback on the visualization sufficiency for model comparison tasks.
We further gathered positive (informal) feedback from two computational linguistic professors on the designed workspace.}
We present insights created for three out of six models introduced in~\autoref{sec:adapter-composition}: the pre-trained BERT, the debiasing adapter for
BERT by Lauscher et al.~\cite{lauscher-etal-2021-sustainable-modular}, and the conll2003 named entity recognizer.
\revision{We plan to extend the study with more participants to quantitatively evaluate the usability of the interface.}


\subsection{Expert Study Setup} 
\revision{The following insights were created collaboratively with two experts in natural language processing tasks. The study was conducted online in the form of a video conference.}
The experts had two main tasks: (1) to investigate models related to bias and (2) to explore the limitations of a named entity recognition model.
The experts further analyzed predictions for sentiment classifiers \textbf{(T4)} as described in~\autoref{sec:prediction}; however, they are not included in the case study description below due to the paper's space considerations.
\revision{The study was concluded with a semi-structured interview about the workspace's usability.}

\noindent\textbf{Data --} The data for the study included the 10 human-interpretable concepts introduced in~\autoref{sec:data-modeling}. The contextualized word embedding representations were extracted from the Yelp dataset~\cite{zhangCharacterlevelConvolutionalNetworks2015}, whereby each word in the concept list was represented by up to 300 contexts.

\noindent\textbf{Tasks --}
For the analysis related to bias detection, the interface provides the debiasing model trained by Lausher et al.~\cite{lauscher-etal-2020-common}.
We use their evaluation results as ground truth to investigate whether the insights can be replicated using our workspace.
In particular, the authors show that the model is effective in
attenuating gender biases according to most of the applied evaluation methods.
However, the results of the Word Embedding Association Test (WEAT)~\cite{doi:10.1126/science.aal4230} are less successful. 
The WEAT test measures the association between two target word sets (e.g., \textit{male pronouns}) and (e.g., \textit{female pronouns}) based on their mean cosine similarity to words from two attribute sets (e.g., \textit{science terms}) and (e.g., \textit{art terms}) that is measured on context-0 (i.e., static~\cite{lauscher-etal-2020-common}) word embeddings.
Lauscher et al. observe that according to the WEAT test, the pre-trained BERT model is insignificantly biased; however, the debiasing adapter does not reduce the bias but instead -- inverts it.
The participants thus received the task to evaluate the particular adapter regarding two specific analysis tasks: (1) to inspect how the embedding space is partitioned for gender-related concepts \textbf{(T1)} and (2) to explore gender-related concept intersections \textbf{(T2)}.

Their second task was to analyze the conll2003 named entity recognizer concerning its learning capabilities of specific named entity categories such as \textit{person names} and \textit{countries}. Their particular analysis tasks were to investigate whether the model partitions the embedding space according to the different categories \textbf{(T1)}, whether there are intersections between the categories \textbf{(T2)}, and whether the model produces `unexpected' associations \textbf{(T3)} between specific named entities.

\vspace{-6pt}
\subsection{Expert Case Studies}\label{sec:case-studies}
In the following, we describe gained insights for the specified tasks. 

\textbf{(Task 1) Bias in Language Models --} 
To gain insights into the gender-related concept representation and their intersections, the participants investigated the Concept Embedding Similarity visualization.
They selected the pre-trained BERT and debiasing models and analyzed the word similarities between different concepts (e.g., \textit{person names} as shown in~\autoref{fig:debiasing}) to \textit{pronouns} that were displayed as anchors in the visualization.
The visualization revealed that in the upper layers (e.g., layer 11) of the pre-trained BERT, context-0 embeddings for \textit{person names} are slightly more similar to \textit{male pronouns} than \textit{female pronouns}, but the difference is insignificant. 
However, in debiasing adapter, most of these \textit{person names} (even \textit{male person names}) are more similar to \textit{female pronouns}.
Similar patterns could be observed for other concepts (e.g., \textit{gender-related stereotypes}, \textit{countries}), which matches the observations by Lauscher et al.~\cite{lauscher-etal-2021-sustainable-modular}. 
It is important to notice that this `bias inversion' is visible only for context-0 embeddings. 
When exploring the relationships between the same concepts computed on contextualized word embeddings (in~\autoref{fig:debiasing}), both Concept Embedding Similarity and Concept Embedding Projection visualizations show that the debiasing adapter was able to eliminate the gender information -- the visualizations show no separation between the \textit{person-name} and \textit{pronoun} concepts. 
However, in the pre-trained BERT, \textit{female person names} are more similar to \textit{female pronouns} and \textit{male person names} are more \begin{wrapfigure}[15]{r}{0.30\textwidth}  
    \centering
    \vspace{-14pt}
    \includegraphics[width=\linewidth]{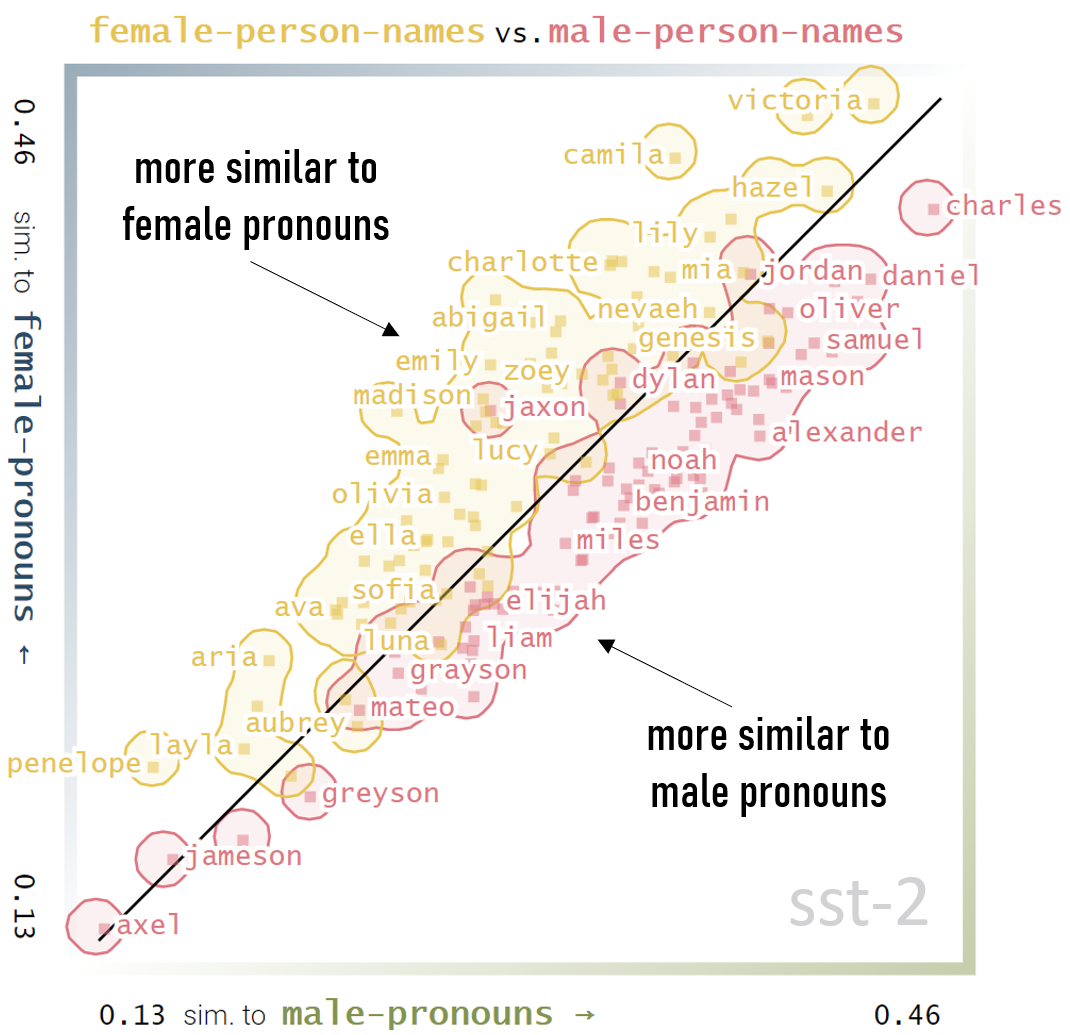}
\end{wrapfigure}similar to \textit{male pronouns}. 
The visualizations reveal that most of the models obtain the gender information from the word's context, and it is not encoded in the word (e.g., \textit{person name}) itself. 
The only exception is the sst-2 sentiment classifier; there, even context-0 embeddings get separated by gender (side figure). 
Different to other adapters, the sst-2 model is trained on phrases extracted from Stanford parse trees rather than full sentences.
Thus, words in isolation that are used to extract the context-0 embeddings present an unnatural input to most of the models~\cite{bommasani-etal-2020-interpreting}; however, the input is less unnatural for the sst-2 model since some of its training instances are one or two words long.

\textbf{(Task 2) Named Entity Recognition --} 
To analyze the learning capabilities of the conll2003 named entity recognizer, the participants explored the Concept Embedding Similarity visualization for the concept \textit{low/high-GDP countries} -- two word lists, each grouping countries with a similar GDP rank according to 2020 statistics. 
As shown in the side \begin{wrapfigure}[15]{l}{0.27\textwidth}  
    \centering
    \vspace{-14pt}
    \includegraphics[width=\linewidth]{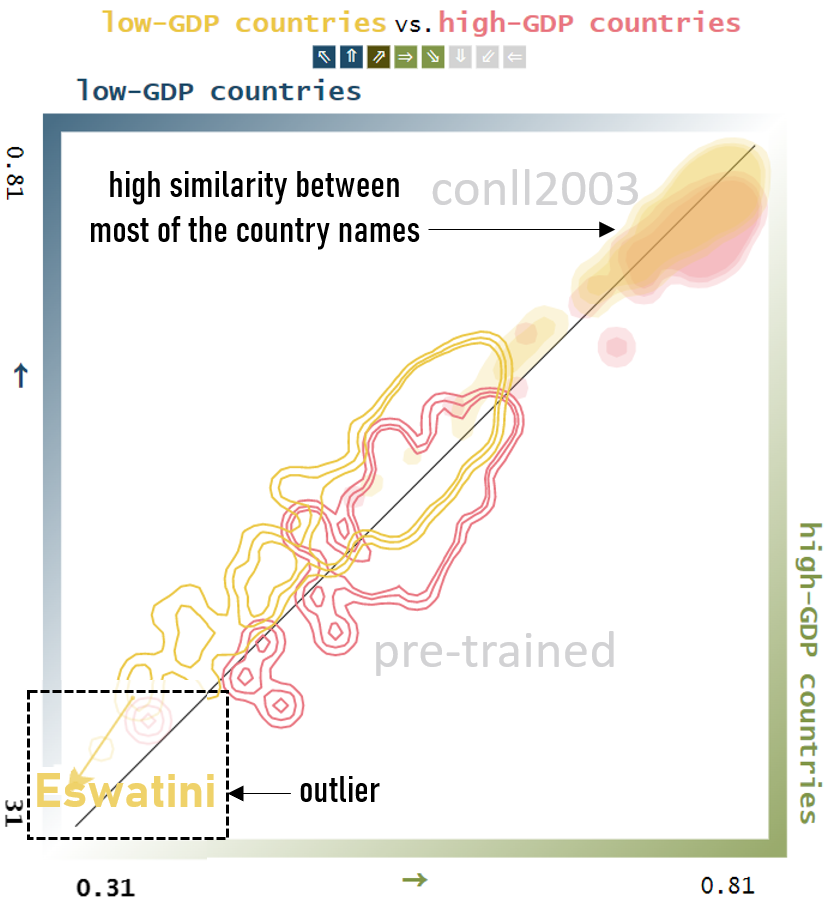}
\end{wrapfigure}figure, the conll2003 model learns that most of the countries are similar without encoding their welfare (see the top-right corner).
By exploring the word positions, one can see that the model does not recognize the country \textit{Eswatini} since its similarity to both \textit{low-GDP} and \textit{high-GDP countries} is low (0.31) compared to other countries that have a similarity of 0.8. 

\begin{figure}[b]
\vspace{-17pt}
    \centering
    \includegraphics[width=\linewidth]{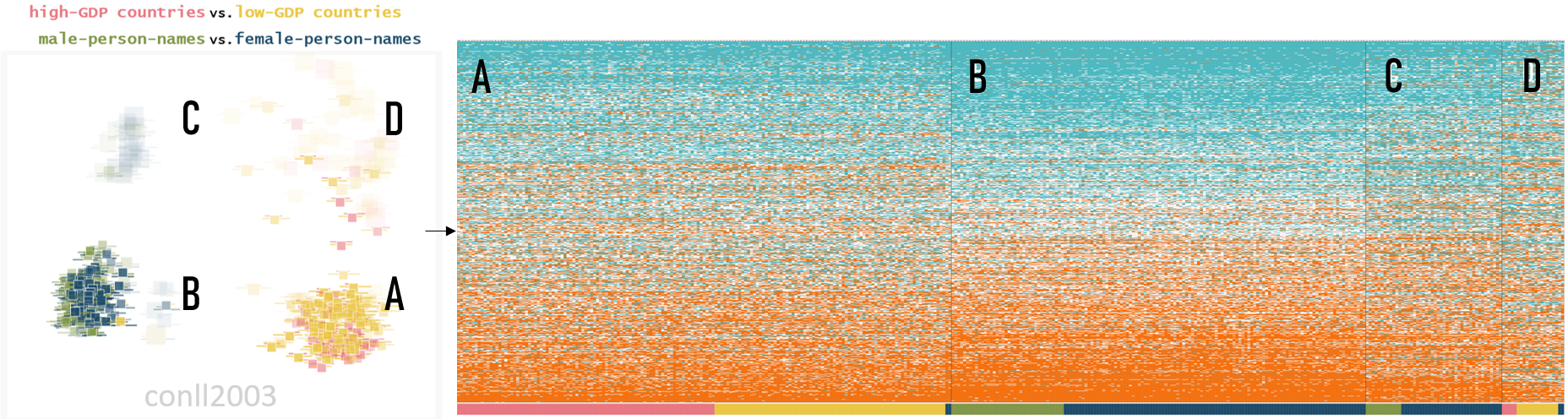}
    \vspace{-15pt}
    \caption{In \textit{Projection Artifact View}, the user can explore embedding vectors aligned as columns in a pixel visualization. 
    We use a bipolar color scale to show vector values (from min \textcolor{bluePixel}{\textbf{blue}} to max \textcolor{orangePixel}{\textbf{orange}}).}
    \label{fig:pixel-vis}
     \vspace{-5pt}
\end{figure}

Next, the participants analyzed the model's distinction between \textit{person names} and \textit{country names} -- a typical task for a named entity recognizer.
The Concept Embedding Projection visualization of the two concepts is shown in~\autoref{fig:projection}.
In the early layers, both models produce similar word neighborhoods and the \textit{person names} and \textit{country names} have a poor separation.
In upper layers (e.g., layer 11 in~\autoref{fig:layer11}), the projection of conll2003 embeddings displays four clusters.
One cluster contains country names (\autoref{fig:pixel-vis} cluster A) and another -- person names (\autoref{fig:pixel-vis} cluster B). The neighborhoods of the two smaller clusters are similar to those in the pre-trained BERT, suggesting that the \textit{conll2003} model did not capture any new properties for these particular words.
By interactively exploring the word neighborhoods, one can observe that one cluster consists of rare person names (e.g., \textit{Nevaeh}), whereas the other contains relatively long country names (e.g., \textit{Trinidad and Tobago}).
Since the visualizations show the context-0 embeddings, the \textit{person names} are not separated by gender.
To investigate whether the four clusters are artifacts generated by the PCA projection, the embeddings values were displayed in the Projection Artifact View. 
\autoref{fig:pixel-vis} shows that the values for embedding vectors within one cluster produce similar patterns, suggesting that the four clusters are not the projection's generated artifacts.
The separation between long and short \textit{country names}, as well as common and rare \textit{person names}, might be a reason of long and rare words not being in the BERT's vocabulary; thus, this might be an artifact of averaging sub-token embedding vectors and must be further investigated.

\vspace{-6pt}
\subsection{Preliminary Expert Feedback}
The experts provided positive feedback concerning the workspace's applicability for model evaluation and comparison tasks.
\revision{They described the interface to be intuitive and easy to use.}
The experts found it useful having the option to choose between different concepts, and in particular--with respect to bias--different ways to quantify it. This allows them to evaluate the models along `different axes', and this is in accordance with works that have shown that bias is manifested in multiple ways. The experts also appreciated the ability to analyze both the representations and the predictions that provide two complementary ways to explain a model: the prediction-based view focuses on the more high level `interface' (i.e., model's predictions) while the representation analysis focuses on its actual working mechanism (i.e., how these predictions are derived). The workspace also demonstrates and makes use of one of the advantages of adapters over other fine-tuning methods -- the fact they are easily integrated into one pre-trained model without having to fine-tune a different model per task.

One important advantage of our workspace was described by the experts as follows. Adapters are usually tested in-domain (e.g., people train for the sentiment task and evaluate on sentiment prediction). The `side-effects' the training has on other aspects are often unaddressed. Thus, it was appreciated that the workspace puts emphasis on evaluating a given adapter according to metrics that are not necessarily related to the main tasks it was trained on.
The interface with its diverse concepts brings another advantage, particularly for the bias evaluation tasks. According to the experts, while certain notions of bias are well studied, the more interesting cases are those which are more subtle and less intuitive or straightforward. The workspace makes it easier to explore the representation space of the models and potentially discover new notions of bias, or more generally, undesired properties of the model in question, as depicted in the~\autoref{sec:case-studies}.
The limitations of the workspace are formulated as research opportunities in the following section.

\vspace{-5pt}
\section{Discussion and Research Opportunities}

In the previous section, we presented how we can use our workspace to gain insights into model specificities. 
During the design and evaluation process, we discovered several opportunities for future research.

\noindent\revision{\textbf{Comparison of Numerous Models --} Currently, our workspace supports the direct comparison of two models at a time. An interesting research challenge would be to display more than two models in the same comparison visualization. While designing our visualizations, we faced challenges in how to select designs that allow visually separate the two models. By displaying more than two models simultaneously, one would need to come up with new visual design alternatives.} 

\noindent\revision{\textbf{Supporting Model Fine-Tuning --} Our work is a step toward effectively comparing adapter models. It is still limited to explorative tasks and, at this point, does not actively suggest which actions to undertake to improve the adapter performances. We see, however, this as a very important direction for future work. The system should provide insights into the models' strengths and limitations and, in an ideal case, also provide hints or suggestions on which steps should be overtaken (e.g., adaptation of the training dataset) to improve the models' performances.}

\noindent\textbf{Visual Explanations Combined with Probing Classifiers --} During our collaboration, the NLP researchers mentioned several potential extensions concerning the functionality of the workspace. Since they commonly train classifiers to investigate concept intersections, they mentioned this as an extension to the visual explanation methods. The two methods used in parallel could increase their trust in the generated insights. 
In particular, if the projection and the classifier produce similar results, it is more likely to be true and less likely to be an artifact of the particular method in use. 

\noindent\textbf{Support for Adapter Training --} 
Currently, our workspace supports the analysis of adapters from the AdapterHub repository. 
The framework, however, supports different adapter composition techniques, such as adapter stacking~\cite{pfeiffer-etal-2020-mad} as well as their fusion~\cite{pfeiffer2020adapterfusion}. 
We plan to extend the workspace in a way that researchers could train new adapters in the interface by applying the different adapter composition methods and directly evaluate their created representation spaces, which, hopefully, would lead to better-performing models for downstream tasks.


\vspace{-5pt}
\section{Conclusion}
We presented a novel visual analytics workspace for the analysis and comparison of LMs that are adapted for different masked language modeling and downstream classification tasks.
The design was motivated by requirements gathered during a literature review and collaboration with NLP researchers.
We introduced three new comparison visualizations: Concept Embedding Similarity, Concept Embedding Projection, and Concept Prediction Similarity that were designed by applying the comparative visualization guidelines by Gleicher~\cite{Gleicher2018ConsiderationsFV}.
\revision{We show the applicability of the workspace through expert case studies, confirm findings from the related work, and generate new insights into adapter learning properties.} 
A demo is available as part of the LingVis framework~\cite{el-assady-etal-2019-lingvis} under: \href{https://adapters.demo.lingvis.io/}{https://adapters.demo.lingvis.io/}.

\vspace{-6pt}
\section*{Acknowledgments}
\vspace{-3pt}
\noindent This paper was funded by the Deutsche Forschungsgemeinschaft (DFG, German Research Foundation)  within projects BU 1806/10-2 ``Questions Visualized'' of the FOR2111, and the ETH AI Center.  


\clearpage
\bibliographystyle{abbrv-doi}

\bibliography{bibliography}

\end{document}